\documentclass[letterpaper, 10 pt, conference]{ieeeconf}  

\IEEEoverridecommandlockouts                              

\usepackage{graphics} 
\usepackage{epsfig} 
\usepackage{times} 
\usepackage{amsmath} 
\usepackage{amssymb}  
\usepackage{multirow,booktabs}
\usepackage{color}
\usepackage[table,HTML]{xcolor}
\usepackage{colortbl}
\usepackage{cite}
\usepackage{bbding}
\usepackage{threeparttable}
\usepackage{stfloats}

\newcommand{\rmnum}[1]{\romannumeral #1}

\title{\LARGE \bf
Collision-Aware and Observation-Aligned Object-Centric\\Scene Reconstruction from Point Cloud
}

\author{Yuxuan Xie$^{1}$, Xuan Yu$^{1}$, Rong Xiong$^{1}$, Yue Wang$^{1}$
\thanks{$^{1}$Yuxuan Xie, Xuan Yu, Rong Xiong, and Yue Wang are with Zhejiang University, Hangzhou, Zhejiang, China. Yue Wang is the corresponding author {\tt\footnotesize wangyue@iipc.zju.edu.cn}}%
}

\begin{document}

\maketitle
\thispagestyle{empty}
\pagestyle{empty}

\begin{abstract}
Object-centric scene reconstruction requires completing partial object observations while preserving metric alignment and avoiding collisions with the surrounding. 
Existing generation-based methods are often image-conditioned and suffer from scale ambiguity and insufficient geometric constraints. 
We propose COOL, a framework for COllision-aware and Observation-aLigned reconstruction.
Based on an object generation model, COOL conditions the generation on instance and background point clouds.
Instance geometry anchors generation in scene coordinates, while background geometry provides local context for scene-consistent completion. 
We further introduce an explicit collision loss and use joint optimization and resampling to reduce collisions during inference. 
Experiments on 3D-Front and Scan2CAD demonstrate strong scene-level fidelity, observation alignment, and collision reduction. 
Moreover, additional studies validate its robustness to mask errors and its applicability to real-world scene replicas.

\end{abstract}

\section{Introduction}



Object-centric scene reconstruction is a task to recover high-fidelity and pose-aligned objects within the scene.
Unlike scene-centric methods that reconstruct single scene entities with incomplete surfaces, object-centric reconstruction focuses on extracting complete and isolated object models\cite{siddiqui2026shaper, yao2025cast}. 
This capability holds promise for constructing digital replicas of real-world scenes and enabling real-to-sim pipelines in robotics.
To achieve this, object-centric reconstruction should complete object geometry independently while preserving metric partial observations and local scene compatibility. This joint objective presents two key challenges, the alignment with observation, and collision with other objects and scene.

Existing methods can be broadly categorized into two classes: regression-based and generation-based.
Regression-based methods\cite{huang2020pf, AdaPoinTr, liu2022towards} learn a one-to-one mapping from partial observations to complete shapes, where the result typically represents an average over all plausible shapes and often becomes overly blurry and smooth. 
Generation-based methods\cite{yao2025cast, ardelean2025gen3dsr, huang2025midi, meng2025scenegen, chen2025sam} offer an alternative by leveraging object generation models \cite{poole2022dreamfusion, liu2023zero, zhang20233dshape2vecset, xiang2025structured} that learn shape priors to synthesize diverse and detailed object geometry. 

Existing generation-based methods mainly condition generation on images and can synthesize high-quality geometry. However, their generated objects are often poorly aligned with the actual observations in the scene.
Some staged pipelines\cite{yao2025cast, ardelean2025gen3dsr} generate in a canonical coordinate system and estimate pose and scale in the scene, causing alignment errors to accumulate across stages. Others\cite{huang2025midi, meng2025scenegen, chen2025sam} predict shape and layout jointly, but remain affected by the scale ambiguity of single-view images and can produce inaccurate object locations and relative sizes.
In addition, independently generated objects may collide with each other or with the surrounding scene. Implicit scene relations encoded in images are often insufficient to prevent these physical collisions. 
Such errors are particularly problematic in robotics, where accurate alignment with sensor observations and collision-free scene geometry are essential for reliable applications.

Depth camera and LiDAR in robotic systems provide accessible metric geometry for scene reconstruction.
Unlike image features, point clouds contain direct metric cues about visible object geometry, scale, and pose.
Thus generation conditioned on point clouds promotes metric alignment and provides a natural basis for checking geometric compatibility with surrounding structures.
Existing point-cloud-conditioned generation methods\cite{chen2025sam, 3D-Fixer}, however, mainly use point clouds to improve object-level generation quality or provide coarse spatial anchors. 
Fully exploiting scene-coordinate metric geometry to model constraints during generation remains unsolved.

\begin{figure}[t]
    \centering
    \includegraphics[width=\linewidth,keepaspectratio]{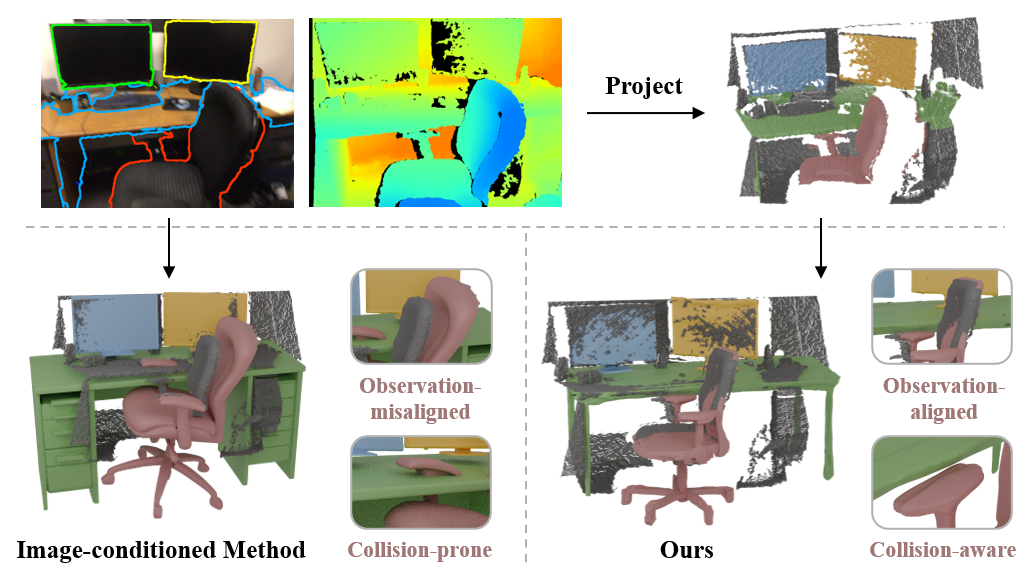}\\
    \vspace{-0.3cm}
    \caption{Image-conditioned generation-based methods for object-centric scene reconstruction produce inaccurate object locations and relative sizes.
    Our method fully exploits scene-coordinate metric geometry to gain an advantage in generating observation-aligned and collision-aware objects.}
    \vspace{-0.3cm}
    \label{fig:first_fig}
\end{figure}

To this end, we propose COOL, a COllision-aware and Observation-aLigned framework for point-cloud-conditioned object-centric scene reconstruction.
The key idea of COOL is to fully exploit scene geometry: in addition to object-level geometry condition, we use the surrounding geometry as a context condition, and introduce generation guidance from explicit geometric collision.
Specifically, building upon a pretrained object generation model \cite{xiang2025structured}, COOL conditions the generation on the observed instance and background point cloud, which are back-projected by masked depth observations.
The instance point cloud conditions generation to preserve metric alignment, while the background point cloud provides local spatial context for scene-consistent completion. 
Both of them play a role in improving observation alignment.
To further resolve collisions, we construct an explicit collision loss between the generated object and the reconstructed surroundings. During inference, COOL jointly guides the generation under this loss and uses resampling to escape local optima. In this way, COOL reduces geometric overlap while preserving fidelity to the observed geometry.


We evaluate COOL on synthetic and real-world scene datasets. Comparative experiments demonstrate its effectiveness in reconstructing observation-aligned and collision-aware object geometries. 
We further evaluate robustness under realistic perception errors, extend the framework to multi-view observations, and conduct ablation studies to analyze the effectiveness of each proposed component.
In summary, our main contributions are as follows:

\begin{itemize}

\item We propose a point-cloud-conditioned object generation framework that reconstructs objects directly in scene coordinates. By using observed instance geometry and local scene geometry as conditions, COOL improves observation fidelity in scale, pose, and scene consistency.
\item We introduce an explicit collision loss between generated objects and reconstructed surroundings, together with a joint optimization and resampling strategy that effectively reduces collisions during inference.
\item COOL achieves better performances in reconstructing scenes from both synthetic and real-world scene datasets, and we validate its robustness under realistic perception errors.

\end{itemize}

\section{Related Work}


\subsection{3D Shape Completion}
Traditional 3D shape completion methods \cite{mitra2006partial, pauly2008discovering} rely on structural regularities, such as the symmetries, to infer missing geometry from partial observations. 
With the increasing availability of large-scale 3D datasets, approaches \cite{huang2020pf, AdaPoinTr} turn to design mapping networks under the supervision of ground truths. These methods aim to learn an optimal one-to-one mapping from partial inputs to complete objects, which often leads to an averaged, blurry solution that lacks fine-grained details. Subsequently, shape completion methods based on generative models \cite{yan2022shapeformer, li2023generalized}, are developed to produce completions from partial inputs, achieving greater generative flexibility. However, they are mainly trained on synthetic datasets, leading to performance degradation when applied to real-world object completion. 



\subsection{3D Object Generation}
3D object generation focuses on generating object geometry and texture from input condition.
Building on the success of 2D generation models \cite{saharia2022photorealistic, rombach2022high}, 3D object generation evolves from optimization methods based on Score Distillation Sampling (SDS) \cite{poole2022dreamfusion} to multi-view consistent reconstruction models \cite{liu2023zero}. However, these approaches struggle to preserve geometric fidelity as they primarily prioritize 2D image quality over explicit 3D structure.
Benefiting from the increase of 3D datasets\cite{fu20213d, deitke2023objaverse} and advances in geometric representations, recent works shift towards training large-scale native 3D generative models \cite{zhang20233dshape2vecset, xiang2025structured}. These approaches typically employ a two-stage pipeline: a Variational Autoencoder (VAE) \cite{king2014auto} compresses 3D geometry into latent space, followed by a latent generative model, which improves both fidelity and efficiency. Nevertheless, these methods remain largely confined to single-object synthesis and assume clean, unobstructed inputs.

\subsection{Object-Centric Scene Reconstruction}

Object-centric scene reconstruction has evolved from retrieval-based CAD alignment \cite{gao2024diffcad} to single-view reconstruction of individual objects and overall scene layout \cite{liu2022towards}, both of which are constrained by limited categories.
Recently, the field has witnessed a paradigm shift toward leveraging generative 3D models to generate objects and reassembling them into the scene.
Most of these methods use image as condition.
Some \cite{ardelean2025gen3dsr, yao2025cast} follow canonical-space object generation with independent pose estimation, suffering from the error accumulation that leads to the misalignment of objects in the scene.
Others \cite{huang2025midi, meng2025scenegen} generate shape together with layout, while they often yield inaccurate relative scales and spatial relations since the single view scale ambiguity.
Besides, current methods\cite{chen2025sam, 3D-Fixer} use point clouds to assist geometric generation. 
SAM3D \cite{chen2025sam} uses a point map as the supplementary of image condition, but lacks explicit geometric constraints.
3D-Fixer \cite{3D-Fixer} leverages point cloud as a spatial
anchor to preserve scene layout, but lacks effective constraints from surrounding geometry.



\section{Method}

\begin{figure*}[t]
    \centering    \includegraphics[width=\linewidth,keepaspectratio]{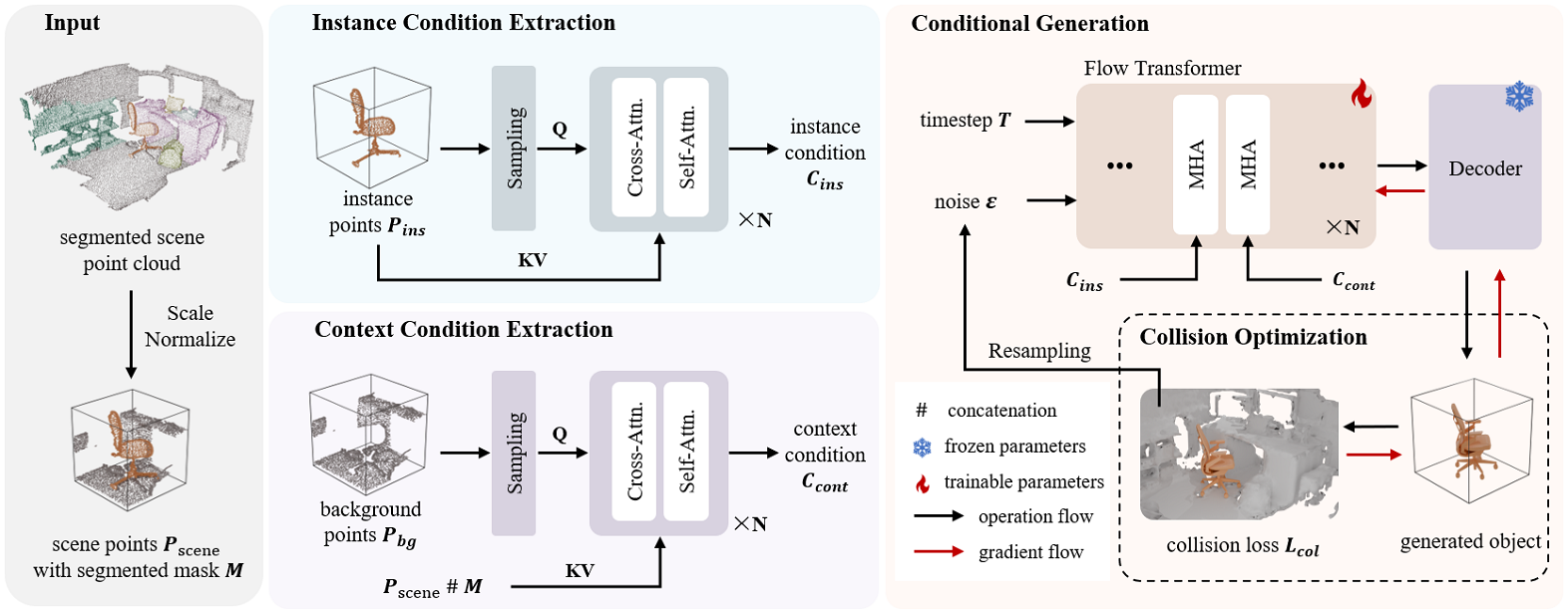}\\
    \vspace{-0.3cm}
    \caption{Overview of COOL. Using segmented and normalized scene point cloud as input, COOL extracts the instance condition and context condition from instance and background point clouds through the condition extraction network. In conditional generation module, these conditions are introduced through MHA layers in the flow transformer. The generated objects are placed back to the reconstructed scene to compute the geometric overlap as a collision loss, optimizing collisions via a joint optimization and resampling strategy during inference.}
    \vspace{-0.3cm}
    \label{fig:pipeline}
\end{figure*}




Given the RGB-D observation $(I,D,P)$ of a scene, where $I$ and $D$ denote the aligned RGB image and depth map and $P$ is the camera projection matrix, together with a target-instance mask $M$, our goal is to reconstruct complete object geometry with observation-aligned scale, pose and shape, which can be directly assembled into a 3D scene.
We back-project valid depth pixels to obtain the scene point cloud $P_{\mathrm{scene}}$, and use mask $M$ to segment it into instance point cloud $P_{\mathrm{ins}}$ and background point cloud $P_{\mathrm{bg}}=P_{\mathrm{scene}}\setminus P_{\mathrm{ins}}$.
COOL models the conditional distribution $v_{\theta}(y\mid P_{\mathrm{ins}},P_{\mathrm{bg}})$, where $y$ is the completed object geometry represented in the original scene coordinate system. 

The remainder of this section details our approach. Based on the 3D object generation model TRELLIS \cite{xiang2025structured} (Sec.\ref{preliminary}), we perform geometry-conditioned instance generation in scene coordinates with instance condition and context condition (Sec.\ref{geometry condition}). Besides, collisions are optimized through a joint optimization and resampling strategy during inference with a collision loss (Sec.\ref{collision optimization}). The training details are introduced in Sec.\ref{training}. Fig.~\ref{fig:pipeline} shows the pipeline of COOL.

\subsection{Preliminary: 3D Object Generation} \label{preliminary}

Our model is built based on TRELLIS \cite{xiang2025structured}, a conditional 3D object generation model, employing rectified flow models to denoise in a structured latent space, where the 3D object is represented as $z = \{(z_i,p_i)\}_{i=1}^{L}$, 
where $p_i \in \{ 0,1,...,N-1 \}^3$ represents the active voxel position of a 3D grid, and $z_i \in \mathbb{R}^C$ is a local latent attached to the voxel.

TRELLIS uses a two-staged pipeline for separate geometry and texture generation, generating the sparse structure first, and then the local latents on it. Both of the stages train a network $v_{\theta}$ based on rectified flow models to move toward the data distribution $x_0$ from $x_t = (1-t)x_0+t \epsilon$, which is the interpolation of data samples $x_0$ and noises $\epsilon$ at timestep $t$. The objective of training the network $v_{\theta}$ is to minimize the conditional flow matching:
\begin{equation}
    \mathbb{E}_{t, x_0, \epsilon}\|v_{\theta}(x_t, t)-(\epsilon-x_0)\|_2^2 .
\end{equation}

During the geometry generation stage, the neural network $\mathcal{G}_S$ generates a low-resolution feature grid $S \in \mathbb{R}^{D \times D \times D \times C_S}$, which can be decoded to a $N$-resolution 3D grid $O$ using a 3D convolutional decoder $\mathcal{D}_S$. The text and image conditions are inserted as the keys and values of the cross attention layers to guide conditional generation.




\subsection{Geometry-Conditioned Instance Generation}
\label{geometry condition}

In this section, we model metric geometry of the object as instance condition to guide generation, which preserves the metric scale and pose of the observation.
Besides, this kind of scene-aligned generation makes scene observation surrounding the object an effective context condition.
The instance and context condition jointly promote observation-aligned generation, as the instance condition anchors the object-level shape, pose, and metric scale, whereas the context condition provides spatial context for scene-level consistency.


{\bf Instance Condition.}
As illustrated in Fig.~\ref{fig:pipeline}, we employ an attention-based point cloud encoder $\mathcal{E}_{ins}$ to derive the instance condition $c_{\text{ins}}$ from the instance partial point cloud $P_{\text{ins}}$.
This process utilizes the sub-sampled instance point cloud $P_{\text{ins}}^s$ to query the original instance point cloud $P_{\text{ins}}$:
\begin{equation}
    c_{\text{ins}} = \operatorname{Attention}(\operatorname{PosEmb}(P_{\text{ins}}^s), \operatorname{PosEmb}(P_{\text{ins}})) ,
\end{equation}
where $\operatorname{PosEmb}(\cdot)$ denotes the positional encoding applied to the 3D coordinates of the point clouds. The attention function $\operatorname{Attention}(\cdot)$ consists of cross-attention layers that aggregate fine-grained geometric details from the original partial points $P_{\text{ins}}$, and self-attention layers that consolidate these details into a coherent, high-level semantic representation. 


{\bf Context Condition.}
With the same structure as $\mathcal{E}_{ins}$, an attention-based encoder $\mathcal{E}_{cont}$ is used to derive the context condition, 
where the sub-sampled background point cloud $P_{\text{bg}}^s$ interacts with the masked scene point cloud $P_{\text{scene}}$ via cross-attention to capture the instance-aware context:
\begin{equation}
    c_{\text{cont}} = \operatorname{Attention}(\operatorname{PosEmb}(P_{\text{bg}}^s), \operatorname{PosEmb}(P_{\text{scene}} \# M)) ,
\end{equation}
where $M$ denotes the binary mask of the instance on the scene point cloud, and $\#$ represents concatenation.

By encoding cross attention between instance and background geometry, context condition incorporates the relationship between instances and their surroundings.

{\bf Generation under Condition.}
%
Based on the shape generation model of text-based TRELLIS, we add two multi-head attention blocks (MHA) to the flow transformer, with one for instance condition $c_{\text{ins}}$ and the other for context condition $c_{\text{cont}}$, as shown in Fig.~\ref{fig:pipeline}.
Within the cross-attention layer, the latent feature $f$ acts as the query to attend to the condition $c$, which serves as the key and value. This operation enables $f$ to incorporate relevant information from $c$:
\begin{equation}
    f^{\prime} = \operatorname{Attention}(f, c), \qquad c\in\{c_{\text{ins}}, c_{\text{cont}}\} .
    \label{eq:cond_attn}
\end{equation}

Guided by the geometry conditions, the generated objects are effectively aligned with the instance partial observations, in terms of geometric shape, metric scale and pose, and scene-level consistency.
By reassembling the independently generated instances, we obtain the reconstructed scene.



\begin{figure}[t]
    \centering
    \includegraphics[width=\linewidth,keepaspectratio]{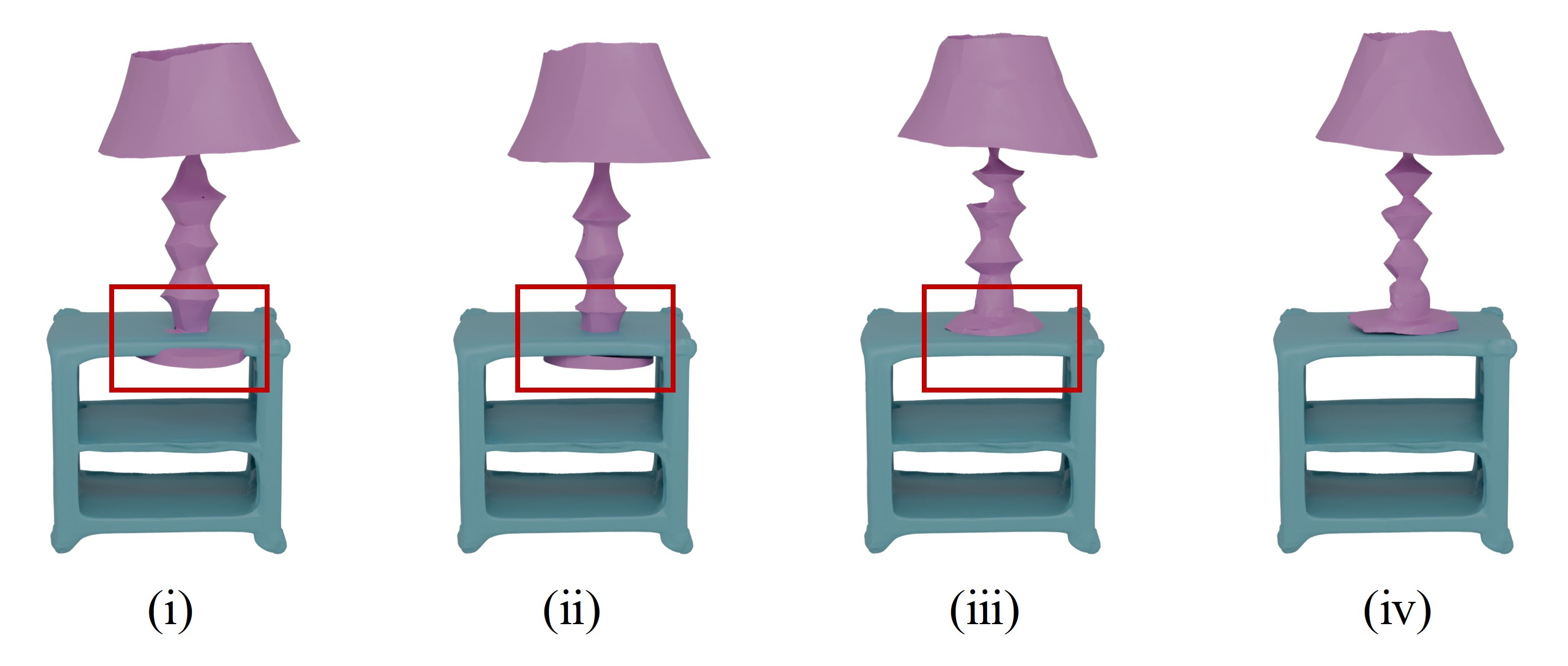}\\
    \vspace{-0.3cm}
    \caption{A case of joint optimization and resampling. (\rmnum{1}) Initial generation with collision. (\rmnum{2}) Gradient-guided optimization converged to a local optima. (\rmnum{3}) Resampling escaping the local optima. (\rmnum{4}) Gradient-guided optimization achieving a collision-free reconstruction.}
    \vspace{-0.3cm}
    \label{fig:col_method}
\end{figure}

\subsection{Collision Optimization} \label{collision optimization}

Despite correct observation alignment and implicit scene context priors, physical collisions are inevitable. To address this, we introduce an explicit collision loss to guide collision-aware denoising trajectory during inference with joint optimization and resampling strategy.

{\bf Collision Loss.}
During inference, we construct an explicit scene-level collision constraint from the observed background geometry. For each target instance, we first remove its pixels from the depth observations using the instance mask, and fuse the remaining depth maps into a background Truncated Signed Distance Function (TSDF) volume\cite{zeng20163dmatch}. 

For each generated sparse-structure grid index $p_i$, we recover its corresponding world coordinate $p_i^w$ through the inverse of the instance normalization transform described in Sec.\ref{training}. 
Let $d_i$ denote the sampled signed distance value at $p_i^w$, and let $m_i \in \{0,1\}$ indicate whether this location is observed in the fused TSDF volume.

Given the predicted sparse structure latent $x_0$, the decoder $D_S$ outputs occupancy logits
\begin{equation}
    o = D_S(x_0), \qquad o_i \in \mathbb{R}.
\end{equation}
A collision occurs when a generated occupied voxel lies inside the observed background, i.e., $o_i > 0$, $d_i < 0$, and $m_i=1$. We define the penetration loss as
\begin{equation}
    \mathcal{L}_{col}
    =
    \sum_i
    \mathbf{1}[m_i=1]\,
    \mathbf{1}[d_i<0]\,
    \mathbf{1}[o_i>0]\,
    (-d_i)_+\,(o_i)_+ .
\end{equation}
This loss penalizes both the penetration depth $-d_i$ and the confidence of generated occupancy $o_i$.

{\bf Gradient and Resampling.}
We use the collision loss to explicitly guide the denoising trajectory during inference without updating the pretrained generator. 
At each denoising step $t$, the flow model predicts a velocity $v_t$, from which the clean sparse structure latent is estimated as $\hat{x}_0 = x_t - tv_t$, where the predicted $\hat{x}_0$ is decoded into occupancy logits and evaluated by the collision loss. Since both the flow model and decoder are frozen, gradients are only used to guide the current sampling trajectory:
\begin{equation}
\begin{aligned}
    g_t &= \nabla_{\hat{x}_0} \mathcal{L}_{col}(D_S(\hat{x}_0)), \\
    \nabla_{v_t}\mathcal{L}_{\mathrm{col}}
    &=
    \left(\frac{\partial \hat{x}_0}{\partial v_t}\right)^{\!\top}
    \nabla_{\hat{x}_0}\mathcal{L}_{\mathrm{col}}
    =
    -tg_t.
\end{aligned}
\end{equation}
We convert this gradient into a correction of the flow velocity. The step size is normalized by the relative magnitudes of the predicted velocity and collision gradient:
\begin{equation}
    \alpha_t =
    \mathrm{clip}
    \left(
    \frac{\rho \|v_t\|_2}
    {t\|g_t\|_2+\epsilon},
    \alpha_{\min}, \alpha_{\max}
    \right),
\end{equation}
where $\rho$ controls the relative strength of collision guidance, $\alpha_{\min}$ and $\alpha_{\max}$ bound the correction magnitude.
A gradient-descent update is performed on the predicted velocity: 
\begin{equation}
\begin{aligned}
\hat{v}_t
&=
v_t-\alpha_t\nabla_{v_t}\mathcal{L}_{\mathrm{col}}
=
v_t+\alpha_t t g_t .
\end{aligned}
\end{equation}
The latent is then updated using the rectified-flow Euler step:
\begin{equation}
    x_{t'}=x_t-(t-t')\hat{v}_t.
\end{equation}
We apply collision guidance during the later denoising stage to preserve the generative prior in early sampling steps.


Because collision-aware optimization is non-convex, a single initial noise may still converge to a poor local solution. Thus we use resampling as a stochastic multi-start strategy. For each instance, we sample $N_s$ independent initial noises
\begin{equation}
    \epsilon^{(k)} \sim \mathcal{N}(0,I), \qquad k=1,\ldots,N_s,
\end{equation}
run the same gradient-guided sampling procedure for each candidate, and select the final result by prioritizing lower collision loss. 
This simple resampling strategy complements gradient-based refinement by exploring multiple plausible regions of the latent space, thereby reducing the chance of being trapped in a collision-prone local optimum. 
Fig.~\ref{fig:col_method} shows a case of joint optimization and resampling.

\begin{table*}[]
\begin{threeparttable} 
\caption{Quantitative comparisons on the 3D-Front dataset and the Scan2CAD dataset}
\centering
\setlength{\tabcolsep}{4.6pt}
\renewcommand\arraystretch{1.2}
\begin{tabular}{lccccc|cc|ccccc|cc}
\toprule
\multirow{2}{*}{\textbf{Method}} & \multicolumn{7}{c}{3D-Front} & \multicolumn{7}{c}{Scan2CAD} \\
\cmidrule(r){2-8} \cmidrule(r){9-15}
& $\mathrm{CD}_{\mathrm{S}}\downarrow$ & 
$\mathrm{FS}_{\mathrm{S}}\uparrow$ &
$\mathrm{CD}_{\mathrm{O}}\downarrow$ &
$\mathrm{FS}_{\mathrm{O}}\uparrow$ &
$\mathrm{IoU}_{\mathrm{B}}\uparrow$ &
$\mathrm{Col}_{\mathrm{O}}\downarrow$ &
$\mathrm{Col}_{\mathrm{S}}\downarrow$ 
& $\mathrm{CD}_{\mathrm{S}}\downarrow$ & 
$\mathrm{FS}_{\mathrm{S}}\uparrow$ &
$\mathrm{CD}_{\mathrm{O}}\downarrow$ &
$\mathrm{FS}_{\mathrm{O}}\uparrow$ &
$\mathrm{IoU}_{\mathrm{B}}\uparrow$ &
$\mathrm{Col}_{\mathrm{O}}\downarrow$ &
$\mathrm{Col}_{\mathrm{S}}\downarrow$ \\
\midrule
AdaPoinTr
    & 0.002 & 0.965 & 0.026 & 0.895 & 0.853 & 0.010 & 0.078
    & 0.018 & 0.841 & 0.062 & 0.532 & 0.597 & 0.329 & 0.023 \\
InstPIFu 
    & 0.065 & 0.633 & 0.088 & 0.507 & 0.269 & 0.460 & 0.503
    & 0.087 & 0.518 & 0.072 & 0.494 & 0.219 & 0.510 & 0.159 \\
MIDI
    & 0.008 & 0.929 & 0.026 & 0.834 & 0.579 & 0.072 & 0.076
    & -- & -- & -- & -- & -- & -- & -- \\
\textbf{Ours} 
   & \textbf{0.001} & \textbf{0.989} & \textbf{0.011} & \textbf{0.904} & \textbf{0.895} & \textbf{0.002} & \textbf{0.003}
   & \textbf{0.015} & \textbf{0.894} & \textbf{0.036} & \textbf{0.745} & \textbf{0.641} & \textbf{0.027} & \textbf{0.000}
   \\
\bottomrule
\end{tabular}
\label{tab:Comparative_baseline}
\end{threeparttable} 
\vspace{-0.3cm}
\end{table*}


\begin{figure*}[t]
    \centering
    \includegraphics[width=\linewidth,keepaspectratio]{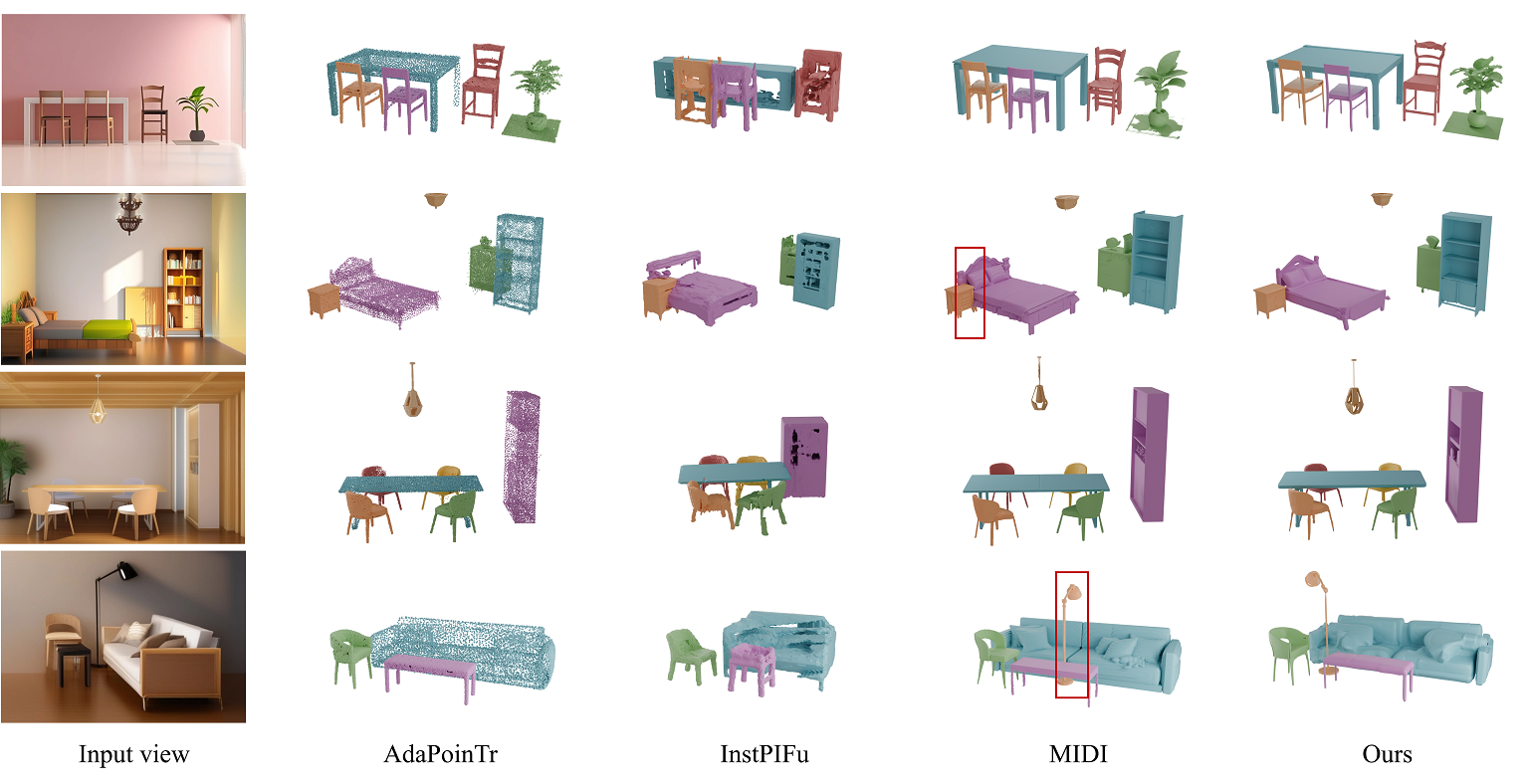}\\
    \vspace{-0.3cm}
    \caption{Qualitative Comparisons on the 3D Front Dataset.}
    \vspace{-0.3cm}
    \label{fig:result_3dfront}
\end{figure*}

\subsection{Training} \label{training}

{\bf Input Preparation.}
We normalize the input instance point cloud using its center $c$ and scale $s$, while applying the same transformation to the associated background points. The generator therefore operates in a normalized coordinate system but retains the relative geometry between the target and its surroundings. The generated object is transformed back using $(c,s)$ before scene assembly.

{\bf Training Strategy.}
During training, we augment the data by randomly merging point clouds from 1 to 5 frames of the same instance.
We finetune the pretrained model of TRELLIS by introducing instance condition and context condition in stages. After the instance condition finetuning stabilizes, we freeze the model parameters and add a new attention module for the context condition, which is then trained separately. This strategy helps better preserve the priors of the pretrained model, leading to more stable and higher‑accuracy results.
\section{Experiments}

\subsection{Setup}

{\bf Datasets.} 
We trained our model on the 3D-Front dataset \cite{fu20213d} and the Scan2CAD dataset \cite{Avetisyan_2019_CVPR}. 
The 3D-Front dataset is a synthetic indoor scene dataset.
We follow the split of prior work \cite{huang2025midi} and use 8K training scenes and 1K test scenes. 
The Scan2CAD dataset is a real-world dataset that matches CAD models to real-world indoor scene scans. We split it into disjoint training and testing sets, and select 10 views for each instance according to visibility and depth quality, constructing 90K training samples with 1.4K scenes.


{\bf Input Protocol.} 
Unless otherwise stated, we use ground-truth instance masks to isolate reconstruction quality. For realistic perception evaluation, we replace them with masks predicted by segmentation models \cite{kirillov2023segment, pr++} while keeping the remaining process unchanged.

{\bf Baselines.}
We compare our method with both regression-based and generation-based methods. The former includes AdaPoinTr \cite{AdaPoinTr} for point cloud completion and InstPIFu \cite{liu2022towards} for feed-forward reconstruction. Generative baselines comprise the multi-stage method Gen3DSR \cite{ardelean2025gen3dsr}, end-to-end methods including image-conditioned MIDI \cite{huang2025midi} and RGBD-conditioned SAM3D \cite{chen2025sam} and 3D-Fixer \cite{3D-Fixer}.
All the methods in comparative study work under a single-view setting. 
COOL uses only geometric observations, while SAM3D and 3D-Fixer additionally use RGB information.

\begin{figure*}[t]
    \centering
    \includegraphics[width=\linewidth,keepaspectratio]{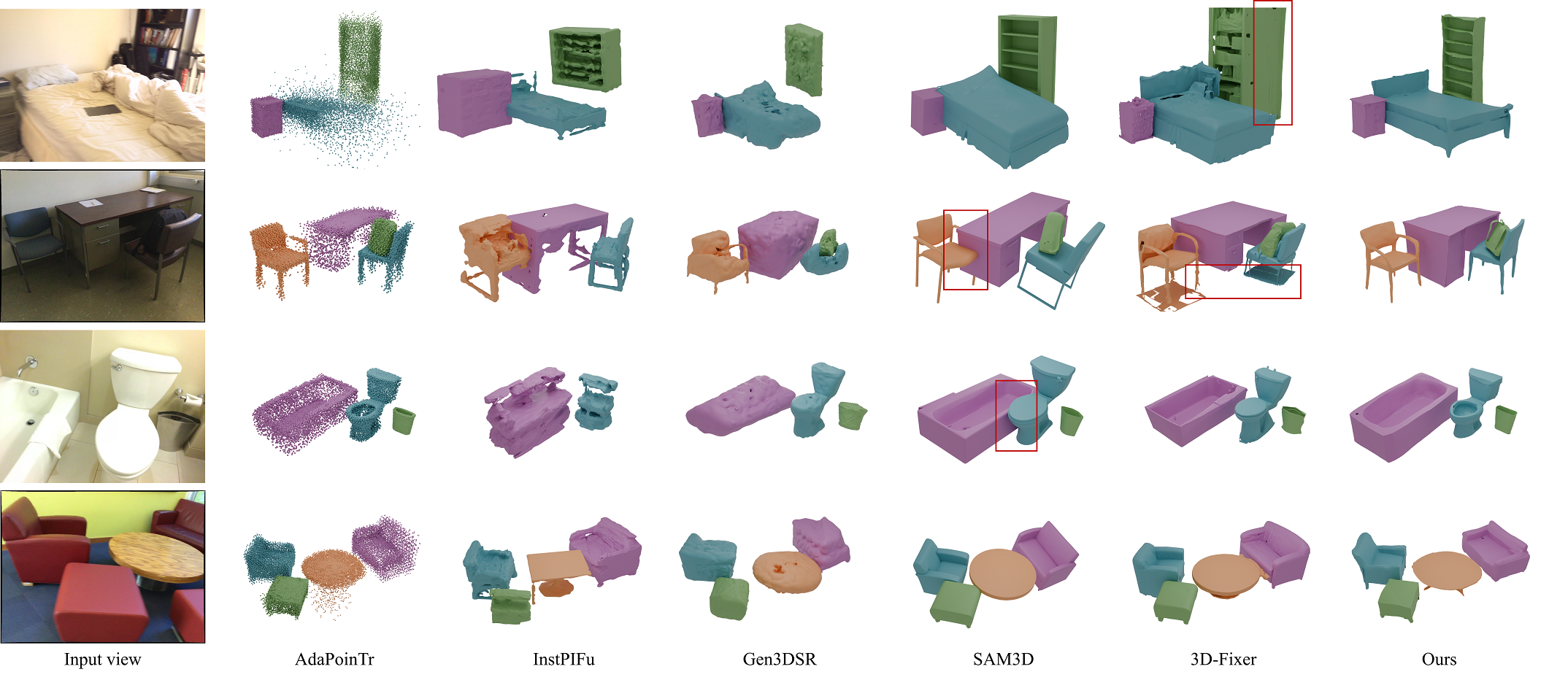}\\
    \vspace{-0.3cm}
    \caption{Qualitative Comparisons on the Scan2CAD Dataset.}
    \vspace{-0.3cm}
    \label{fig:result_scan2cad}
\end{figure*}

{\bf Evaluation Metrics.}
We compute both object-level and scene-level Chamfer Distance (CD) and F-Score (FS) to evaluate the quality of geometric reconstruction. The Volumetric Intersection over Union (IoU) of objects is used to assess the accuracy of spatial layout in generated scenes. Additionally, we employ the inter-object collision rate ($\mathrm{Col}_{\mathrm{O}}$) and object-scene collision rate ($\mathrm{Col}_{\mathrm{S}}$) to report the collision situation.

{\bf Implementation.}
Before condition extraction, input point clouds are sampled by Farthest Point Sampling (FPS) to 512. The condition extraction module includes 8 attention blocks and the dimension of query is 768. 
Our method is based on the pretrained base text-to-3D model of TRELLIS\cite{xiang2025structured} and we follow the same sampling schedule and strategy for generation.
For collision-aware inference, we use 25 rectified-flow sampling steps and collision guidance is activated during the last 13 sampling steps. We set the relative guidance strength to $\rho=0.10$, with $\alpha_{\min}=0.1$ and $\alpha_{\max}=10.0$. For resampling, we generate $N_s=4$ candidates for each instance and select the result with the lowest collision loss.

\begin{table}[]
\caption{Quantitative comparisons on the Scan2CAD dataset with zero-shot methods}
\vspace{-0.2cm}
\centering
\setlength{\tabcolsep}{3pt}
\renewcommand\arraystretch{1.2}
\begin{tabular}{lccccc|cc}
\toprule
Method  & 
$\mathrm{CD}_{\mathrm{S}}\downarrow$ & 
$\mathrm{FS}_{\mathrm{S}}\uparrow$ &
$\mathrm{CD}_{\mathrm{O}}\downarrow$ &
$\mathrm{FS}_{\mathrm{O}}\uparrow$ &
$\mathrm{IoU}_{\mathrm{B}}\uparrow$ &
$\mathrm{Col}_{\mathrm{O}}\downarrow$ &
$\mathrm{Col}_{\mathrm{S}}\downarrow$ \\
\midrule
Gen3DSR
    & 0.050 & 0.655 & 0.085 & 0.443 & 0.398 & 0.291 & 0.093 \\
SAM3D
    & 0.032 & 0.747 & \textbf{0.040} & \textbf{0.691} & 0.382 & 0.322 & 0.050 \\
3D-Fixer
    & 0.042 & 0.791 & 0.059 & 0.595 & 0.464 & 0.286 & 0.067 \\
Ours(zs)
    & \textbf{0.029} & \textbf{0.841} & 0.068 & 0.587 & \textbf{0.549} & \textbf{0.041} & \textbf{0.000} \\
\bottomrule
\end{tabular}
\label{tab:zero-shot comparisons}
\vspace{-0.5cm}
\end{table}


\begin{table*}[b]
\vspace{-0.3cm}
\caption{Ablation Studies on the Scan2CAD dataset}
\vspace{-0.2cm}
\centering
\setlength{\tabcolsep}{5.6pt}
\renewcommand\arraystretch{1.2}
\begin{tabular}{c|c|c|ccccc|cc|c}
\toprule
C.C. &  C.L. & R.S. &
$\mathrm{CD}_{\mathrm{S}}\downarrow$ & 
$\mathrm{FS}_{\mathrm{S}}\uparrow$ &
$\mathrm{CD}_{\mathrm{O}}\downarrow$ &
$\mathrm{FS}_{\mathrm{O}}\uparrow$ &
$\mathrm{IoU}_{\mathrm{B}}\uparrow$ &
$\mathrm{Col}_{\mathrm{O}}\downarrow$ &
$\mathrm{Col}_{\mathrm{S}}\downarrow$ &
$\mathrm{t}_\mathrm{infer}$$\downarrow$\\
\midrule
\XSolidBrush & \XSolidBrush & \XSolidBrush
    & 0.019 & 0.876 & 0.040 & 0.710 & 0.618 & 0.245 & 0.016 & 3.2s \\
\Checkmark & \XSolidBrush & \XSolidBrush
    & \textbf{0.014} & \textbf{0.896} & \underline{0.037} & \underline{0.744} & 0.624 & 0.203 & 0.015 & 3.4s \\
\Checkmark & \Checkmark & \XSolidBrush
    & 0.016 & 0.891 & 0.038 & 0.742 & \underline{0.633} & 0.095 & \underline{0.003} & 3.9s \\
\Checkmark & \XSolidBrush & \Checkmark
    & 0.018 & 0.885 & 0.038 & 0.740 & 0.622 & \underline{0.089} & 0.009 & 5.9s \\
\Checkmark & \Checkmark & \Checkmark
    & \underline{0.015} & \underline{0.894} & \textbf{0.036} & \textbf{0.745} & \textbf{0.641} & \textbf{0.027} & \textbf{0.000} & 7.8s \\
\bottomrule
\end{tabular}
\label{tab:Ablation_context_condition}
    \begin{tablenotes} 
		\item \qquad \qquad \qquad \qquad C.C. denotes context condition, C.L. denotes collision-loss guidance, R.S. denotes resampling.
    \end{tablenotes} 
\vspace{-0.3cm}
\end{table*}

\subsection{Comparative Study}
We evaluate COOL under two complementary protocols. Tab. \ref{tab:Comparative_baseline} reports in-domain comparisons with baselines available training on the corresponding dataset. Each method is evaluated on the test split after training on the training split of 3D-Front or Scan2CAD, respectively. 
In contrast, Tab. \ref{tab:zero-shot comparisons} reports zero-shot transfer to Scan2CAD for methods that are generalizable and only available as pretrained checkpoints. For a comparable setting, Ours(zs) is trained on 3D-Front only and directly evaluated on the Scan2CAD test split.

{\bf Quantitative Results.}
In Tab. \ref{tab:Comparative_baseline}, COOL achieves superior performance on both 3D-Front and Scan2CAD. It consistently improves object-level reconstruction quality, scene-level fidelity, and collision rates.
Tab. \ref{tab:zero-shot comparisons} further evaluates cross-dataset transfer to Scan2CAD without using its training data. 
SAM3D produces strong object-level geometry under its official zero-shot setting, obtaining the best object-level CD and F-score.
Although COOL(zs) obtains lower object-level reconstruction quality than SAM3D, it achieves better scene-level reconstruction quality and collision mitigation. These results indicate that, while zero-shot object shape quality remains limited by the training data, the proposed instance and context geometry conditions effectively preserve scene-coordinate scale and pose alignment, producing spatially consistent and physically compatible scene reconstructions than the compared zero-shot methods.

{\bf Qualitative Results.}
Fig. ~\ref{fig:result_3dfront} and Fig. ~\ref{fig:result_scan2cad} present the qualitative comparisons on the two datasets.
For object-level quality, generation-based approaches generally outperform regression-based baselines. 
AdaPoinTr excels in learning dataset-specific shape correspondences, thus performs well on 3D-Front, whose training and test sets share the CAD library, but its quality visibly degrades on Scan2CAD.

Generation quality, however, does not ensure observation-aligned reconstruction. 
Gen3DSR generates each object through multiple stages, where errors in image completion and pose estimation accumulate.
MIDI estimates spatial arrangement primarily from image cues, resulting in inaccurate relative scales and object locations.
SAM3D uses RGB-D as condition, but its pointmap acts as a generative condition rather than an explicit observation-fitting constraint. Thus its generated pose and scale fail to tightly align with the observation, leading to lower volumetric IoU and higher collision rates.
3D-Fixer uses partial geometry to anchor object scale and pose, but does not constrain the generation against surrounding scene geometry, insufficient to ensure physically compatible placement.

\begin{figure}[t]
    \centering
    \includegraphics[width=\linewidth,keepaspectratio]{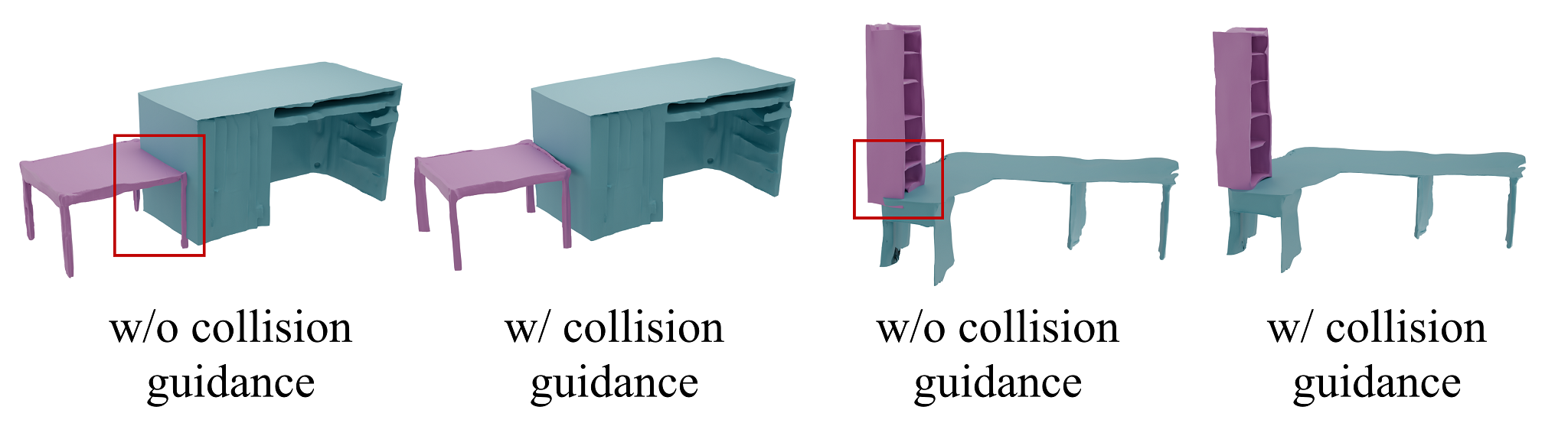}\\
    \vspace{-0.5cm}
    \caption{Qualitative Results on Collision Loss Guidance.}
    \vspace{-0.7cm}
    \label{fig:collision}
\end{figure}

\subsection{Ablation Study} \label{study on collision-aware strategy}
We conduct ablation studies on the Scan2CAD dataset to validate the effectiveness of each component.
Tab. \ref{tab:Ablation_context_condition} reports reconstruction metrics, collision rates and inference time.

{\bf Context Condition.}
The first two rows of Tab. \ref{tab:Ablation_context_condition} evaluate the effect of the context condition. Adding the context condition improves both reconstruction and spatial consistency, improving scene and object level reconstruction metrics and reducing collisions.
However, the collision rates remain high. Therefore, context condition improves scene-consistent generation but cannot explicitly enforce collision-free geometry.

{\bf Collision Loss.}
The third row evaluates collision-guided denoising using one sample.
Introducing collision-loss guidance reduces the object-object collision rate from 0.203 to 0.095 and the object-scene collision rate from 0.015 to 0.003.
Meanwhile, the object-level and spatial-layout metrics remain close to those of the unoptimized result.
This indicates that collision-loss guidance can modify the denoising trajectory to resolve geometric intersections without compromising the plausibility and observation consistency.
Fig.~\ref{fig:collision} shows cases of collision-guided optimization.

{\bf Joint Optimization and Resampling.}
The last two rows evaluate the effect of resampling with four samples.
The fourth row confirms that stochastic sampling can occasionally discover lower-collision solutions while it does not reliably eliminate collisions.
When collision-loss-guided optimization is applied to all four samples, collision rates are further reduced.
These show that resampling and  optimization are complementary: optimization steers each sample toward lower-collision geometry, while resampling explores multiple plausible solutions.
The object-level inference time shows that using four samples provides a practical balance between collision reduction and computational cost.

\begin{table}[t]
\caption{Study on Robustness under Perception Noise}
\vspace{-0.2cm}
\centering
\setlength{\tabcolsep}{5.6pt}
\renewcommand\arraystretch{1.2}
\begin{tabular}{lccccc}
\toprule
Mask & 
$\mathrm{CD}_{\mathrm{S}}\downarrow$ & 
$\mathrm{FS}_{\mathrm{S}}\uparrow$ &
$\mathrm{CD}_{\mathrm{O}}\downarrow$ &
$\mathrm{FS}_{\mathrm{O}}\uparrow$ &
$\mathrm{IoU}_{\mathrm{B}}\uparrow$ \\
\midrule
SAM \cite{kirillov2023segment}
    & 0.010 & 0.985 & 0.024 & 0.810 & 0.727  \\
PanoRecon \cite{pr++}
    & 0.002 & 0.994 & 0.016 & 0.810 & 0.766  \\
GT
    & \textbf{0.001} & \textbf{0.995} & \textbf{0.015} & \textbf{0.816} & \textbf{0.767}  \\
\bottomrule
\end{tabular}
\label{tab:realistic perception}
\end{table}

\begin{figure}[t]
    \centering
    \includegraphics[width=\linewidth,keepaspectratio]{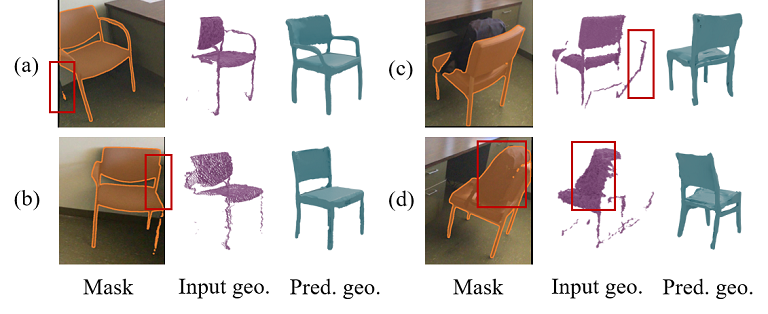}\\
    \vspace{-0.3cm}
    \caption{Qualitative Results under Perception Noise.}
    \vspace{-0.3cm}
    \label{fig:noise}
\end{figure}

\subsection{Study on Robustness under Perception Noise}
We evaluate COOL under two realistic mask sources: SAM, a 2D segmentation model\cite{kirillov2023segment}, and PanoRecon, a panoptic reconstruction pipeline\cite{pr++}.
Tab. \ref{tab:realistic perception} reports the results on 4 scenes in Scan2CAD. 
Compared with GT masks, SAM masks cause a performance decrease mainly caused by missed detection and incorrect segmentation, which can be substantially solved by PanoRecon, since it uses 3D consistency to correct these types of segmentation errors.

Fig. \ref{fig:noise} provides qualitative analysis of different error types under SAM masks. When the mask omits part of the object, COOL can usually recover object structure using generative prior (Fig. \ref{fig:noise}(a)), while it may cause incomplete details (Fig. \ref{fig:noise}(b)).
For point-cloud noise, depth-discontinuous outliers, commonly caused by RGB-depth misalignment near object boundaries, are largely suppressed by the point-cloud condition encoder (Fig. \ref{fig:noise}(c)). In contrast, mask errors can introduce depth-continuous points from nearby objects, which have a stronger influence on the generation (Fig. \ref{fig:noise}(d)).
This result shows that COOL is robust to partial observations and isolated depth outliers, while geometric noise from incorrect masks remains a challenging failure mode.

\begin{figure}[b]
    \vspace{-0.3cm}
    \centering
    \includegraphics[width=\linewidth,keepaspectratio]{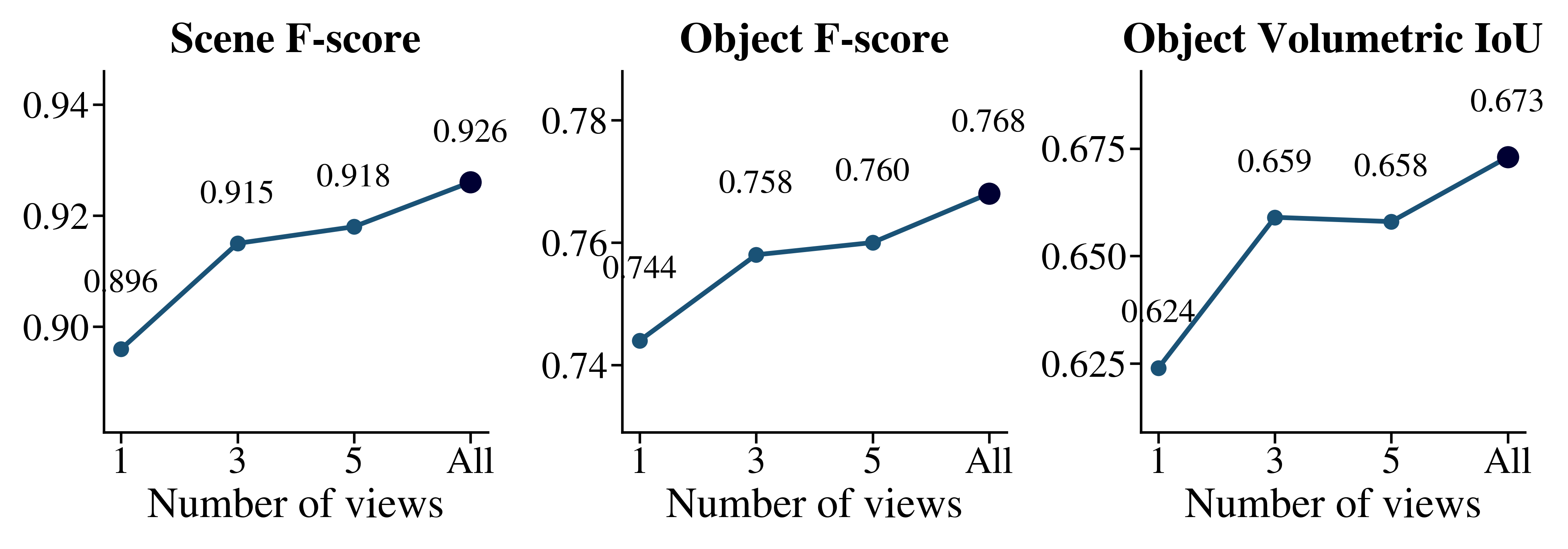}\\
    \vspace{-0.3cm}
    \caption{Study on Multi-view Condition on the Scan2CAD Dataset.}
    \vspace{-0.3cm}
    \label{fig:multi-view}
\end{figure}


\subsection{Study on Multi-view Condition}
COOL natively supports multi-view point cloud inputs. 
We conducted experiments on Scan2CAD with 1, 3, 5-frame inputs, and point clouds sampled from reconstructed scene meshes (denoted as "all"). 
As shown in Fig. \ref{fig:multi-view}, performance improves as more views are integrated, which confirms that richer observations enhance both local object fidelity and global spatial consistency.

\begin{figure*}[t]
    \centering
    \includegraphics[width=\linewidth,keepaspectratio]{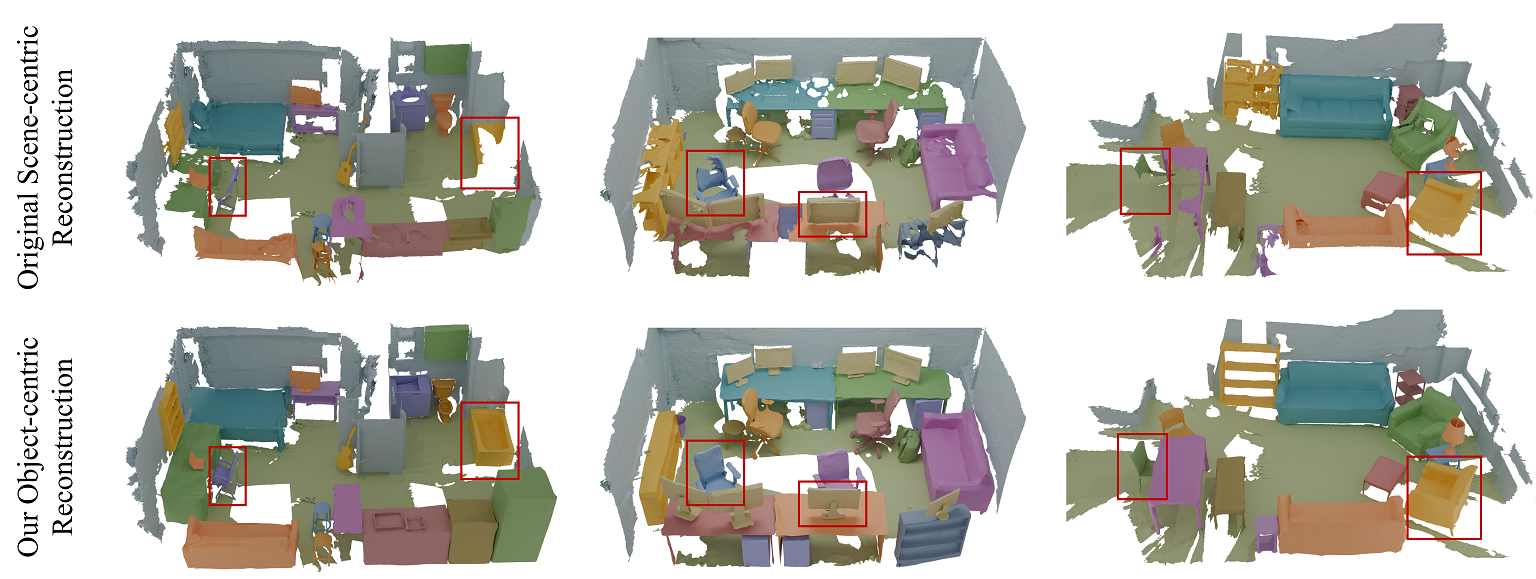}\\
    \caption{Cases on Geometric Replicas of Real-World Scenes.}
    \vspace{-0.3cm}
    \label{fig:cases}
\end{figure*}

\subsection{Case Study}
Fig. \ref{fig:cases} shows COOL$^\prime$s performance in constructing geometric replicas for real-world scenes on ScanNet. COOL reconstructs and integrates complete object geometries into the scene to replace the incomplete objects in scene-centric reconstruction, showing its real-to-sim potential.

\section{Conclusion}

We propose COOL, a framework for object-centric scene reconstruction from geometric observations.
Geometry conditions promote observation-aligned and scene-consistent generation.
During inference, collision-guided optimization together with resampling, further reduces collisions.
Experimental results validate its efficacy and show its applicability in reconstructing real-world replicas for robotics.



{\bf Limitation.}
Compared with image-conditioned generators using richer visual cues, geometry-conditioned generation may lose fine-scale shape details. 
Limited computing resources restrict current training to relatively small datasets, so the generalization ability has not been fully explored.
\bibliographystyle{IEEEtran}
\bibliography{IEEEabrv,bibliography}

\begin{thebibliography}{10}
\providecommand{\url}[1]{#1}
\csname url@samestyle\endcsname
\providecommand{\newblock}{\relax}
\providecommand{\bibinfo}[2]{#2}
\providecommand{\BIBentrySTDinterwordspacing}{\spaceskip=0pt\relax}
\providecommand{\BIBentryALTinterwordstretchfactor}{4}
\providecommand{\BIBentryALTinterwordspacing}{\spaceskip=\fontdimen2\font plus
\BIBentryALTinterwordstretchfactor\fontdimen3\font minus \fontdimen4\font\relax}
\providecommand{\BIBforeignlanguage}[2]{{%
\expandafter\ifx\csname l@#1\endcsname\relax
\typeout{** WARNING: IEEEtran.bst: No hyphenation pattern has been}%
\typeout{** loaded for the language `#1'. Using the pattern for}%
\typeout{** the default language instead.}%
\else
\language=\csname l@#1\endcsname
\fi
#2}}
\providecommand{\BIBdecl}{\relax}
\BIBdecl

\bibitem{siddiqui2026shaper}
Y.~Siddiqui, D.~Frost, S.~Aroudj, A.~Avetisyan, H.~Howard-Jenkins, D.~DeTone, P.~Moulon, Q.~Wu, Z.~Li, J.~Straub, R.~Newcombe, and J.~Engel, ``Shaper: Robust conditional 3d shape generation from casual captures,'' in \emph{Proceedings of the IEEE/CVF Conference on Computer Vision and Pattern Recognition (CVPR)}, June 2026, pp. 27\,157--27\,168.

\bibitem{yao2025cast}
K.~Yao, L.~Zhang, X.~Yan, Y.~Zeng, Q.~Zhang, L.~Xu, W.~Yang, J.~Gu, and J.~Yu, ``Cast: Component-aligned 3d scene reconstruction from an rgb image,'' \emph{ACM Transactions on Graphics (TOG)}, vol.~44, no.~4, pp. 1--19, 2025.

\bibitem{huang2020pf}
Z.~Huang, Y.~Yu, J.~Xu, F.~Ni, and X.~Le, ``Pf-net: Point fractal network for 3d point cloud completion,'' in \emph{2020 IEEE/CVF Conference on Computer Vision and Pattern Recognition}, 2020, pp. 7659--7667.

\bibitem{AdaPoinTr}
X.~Yu, Y.~Rao, Z.~Wang, J.~Lu, and J.~Zhou, ``Adapointr: Diverse point cloud completion with adaptive geometry-aware transformers,'' \emph{IEEE Transactions on Pattern Analysis and Machine Intelligence}, vol.~45, no.~12, pp. 14\,114--14\,130, 2023.

\bibitem{liu2022towards}
H.~Liu, Y.~Zheng, G.~Chen, S.~Cui, and X.~Han, ``Towards high-fidelity single-view holistic reconstruction of indoor scenes,'' in \emph{European Conference on Computer Vision}, 2022, pp. 429--446.

\bibitem{ardelean2025gen3dsr}
A.~Ardelean, M.~{\"O}zer, and B.~Egger, ``Gen3dsr: Generalizable 3d scene reconstruction via divide and conquer from a single view,'' in \emph{2025 International Conference on 3D Vision (3DV)}, 2025, pp. 616--626.

\bibitem{huang2025midi}
Z.~Huang, Y.-C. Guo, X.~An, Y.~Yang, Y.~Li, Z.-X. Zou, D.~Liang, X.~Liu, Y.-P. Cao, and L.~Sheng, ``Midi: Multi-instance diffusion for single image to 3d scene generation,'' in \emph{2025 IEEE/CVF Conference on Computer Vision and Pattern Recognition (CVPR)}, 2025, pp. 23\,646--23\,657.

\bibitem{meng2025scenegen}
Y.~Meng, H.~Wu, Y.~Zhang, and W.~Xie, ``Scenegen: Single-image 3d scene generation in one feedforward pass,'' in \emph{2026 International Conference on 3D Vision (3DV)}, 2026, pp. 543--553.

\bibitem{chen2025sam}
X.~Chen, F.-J. Chu, P.~Gleize, K.~J. Liang, A.~Sax, H.~Tang, W.~Wang, M.~Guo, T.~Hardin, X.~Li \emph{et~al.}, ``Sam 3d: 3dfy anything in images,'' in \emph{Proceedings of the IEEE/CVF Conference on Computer Vision and Pattern Recognition}, 2026, pp. 7220--7232.

\bibitem{poole2022dreamfusion}
B.~Poole, A.~Jain, J.~T. Barron, and B.~Mildenhall, ``Dreamfusion: Text-to-3d using 2d diffusion,'' \emph{arXiv preprint arXiv:2209.14988}, 2022.

\bibitem{liu2023zero}
R.~Liu, R.~Wu, B.~Van~Hoorick, P.~Tokmakov, S.~Zakharov, and C.~Vondrick, ``Zero-1-to-3: Zero-shot one image to 3d object,'' in \emph{2023 IEEE/CVF International Conference on Computer Vision (ICCV)}, 2023, pp. 9264--9275.

\bibitem{zhang20233dshape2vecset}
B.~Zhang, J.~Tang, M.~Niessner, and P.~Wonka, ``3dshape2vecset: A 3d shape representation for neural fields and generative diffusion models,'' \emph{ACM Transactions On Graphics (TOG)}, vol.~42, no.~4, pp. 1--16, 2023.

\bibitem{xiang2025structured}
J.~Xiang, Z.~Lv, S.~Xu, Y.~Deng, R.~Wang, B.~Zhang, D.~Chen, X.~Tong, and J.~Yang, ``Structured 3d latents for scalable and versatile 3d generation,'' in \emph{2025 IEEE/CVF Conference on Computer Vision and Pattern Recognition (CVPR)}, 2025, pp. 21\,469--21\,480.

\bibitem{3D-Fixer}
Z.-X. Yin, L.~Liu, X.~Wang, W.~Sui, Z.~Su, J.~Yang, and J.~Xie, ``3d-fixer: Coarse-to-fine in-place completion for 3d scenes from a single image,'' in \emph{Proceedings of the IEEE/CVF Conference on Computer Vision and Pattern Recognition (CVPR)}, June 2026, pp. 12\,753--12\,763.

\bibitem{mitra2006partial}
N.~J. Mitra, L.~J. Guibas, and M.~Pauly, ``Partial and approximate symmetry detection for 3d geometry,'' \emph{ACM Transactions on Graphics (ToG)}, vol.~25, no.~3, pp. 560--568, 2006.

\bibitem{pauly2008discovering}
M.~Pauly, N.~J. Mitra, J.~Wallner, H.~Pottmann, and L.~J. Guibas, ``Discovering structural regularity in 3d geometry,'' in \emph{ACM SIGGRAPH 2008 papers}, 2008, pp. 1--11.

\bibitem{yan2022shapeformer}
X.~Yan, L.~Lin, N.~J. Mitra, D.~Lischinski, D.~Cohen-Or, and H.~Huang, ``Shapeformer: Transformer-based shape completion via sparse representation,'' in \emph{2022 IEEE/CVF Conference on Computer Vision and Pattern Recognition (CVPR)}, 2022, pp. 6229--6239.

\bibitem{li2023generalized}
Y.~Li, Y.~Dou, X.~Chen, B.~Ni, Y.~Sun, Y.~Liu, and F.~Wang, ``Generalized deep 3d shape prior via part-discretized diffusion process,'' in \emph{2023 IEEE/CVF Conference on Computer Vision and Pattern Recognition (CVPR)}, 2023, pp. 16\,784--16\,794.

\bibitem{saharia2022photorealistic}
C.~Saharia, W.~Chan, S.~Saxena, L.~Li, J.~Whang, E.~L. Denton, K.~Ghasemipour, R.~Gontijo~Lopes, B.~Karagol~Ayan, T.~Salimans \emph{et~al.}, ``Photorealistic text-to-image diffusion models with deep language understanding,'' \emph{Advances in neural information processing systems}, vol.~35, pp. 36\,479--36\,494, 2022.

\bibitem{rombach2022high}
R.~Rombach, A.~Blattmann, D.~Lorenz, P.~Esser, and B.~Ommer, ``High-resolution image synthesis with latent diffusion models,'' in \emph{2022 IEEE/CVF Conference on Computer Vision and Pattern Recognition (CVPR)}, 2022, pp. 10\,674--10\,685.

\bibitem{fu20213d}
H.~Fu, B.~Cai, L.~Gao, L.-X. Zhang, J.~Wang, C.~Li, Q.~Zeng, C.~Sun, R.~Jia, B.~Zhao, and H.~Zhang, ``3d-front: 3d furnished rooms with layouts and semantics,'' in \emph{2021 IEEE/CVF International Conference on Computer Vision (ICCV)}, 2021, pp. 10\,913--10\,922.

\bibitem{deitke2023objaverse}
M.~Deitke, D.~Schwenk, J.~Salvador, L.~Weihs, O.~Michel, E.~VanderBilt, L.~Schmidt, K.~Ehsanit, A.~Kembhavi, and A.~Farhadi, ``Objaverse: A universe of annotated 3d objects,'' in \emph{2023 IEEE/CVF Conference on Computer Vision and Pattern Recognition (CVPR)}, 2023, pp. 13\,142--13\,153.

\bibitem{king2014auto}
D.~King and M.~Welling, ``Auto-encoding variational bayes,'' in \emph{Proceedings of the 2nd International Conference on Learning Representations (ICLR), Banff, AB, Canada}, 2014, pp. 14--16.

\bibitem{gao2024diffcad}
D.~Gao, D.~Rozenberszki, S.~Leutenegger, and A.~Dai, ``Diffcad: Weakly-supervised probabilistic cad model retrieval and alignment from an rgb image,'' \emph{ACM Transactions on Graphics (TOG)}, vol.~43, no.~4, pp. 1--15, 2024.

\bibitem{zeng20163dmatch}
A.~Zeng, S.~Song, M.~Nießner, M.~Fisher, J.~Xiao, and T.~Funkhouser, ``3dmatch: Learning local geometric descriptors from rgb-d reconstructions,'' in \emph{2017 IEEE Conference on Computer Vision and Pattern Recognition (CVPR)}, 2017, pp. 199--208.

\bibitem{Avetisyan_2019_CVPR}
A.~Avetisyan, M.~Dahnert, A.~Dai, M.~Savva, A.~X. Chang, and M.~Nießner, ``Scan2cad: Learning cad model alignment in rgb-d scans,'' in \emph{2019 IEEE/CVF Conference on Computer Vision and Pattern Recognition (CVPR)}, 2019, pp. 2609--2618.

\bibitem{kirillov2023segment}
A.~Kirillov, E.~Mintun, N.~Ravi, H.~Mao, C.~Rolland, L.~Gustafson, T.~Xiao, S.~Whitehead, A.~C. Berg, W.-Y. Lo \emph{et~al.}, ``Segment anything,'' in \emph{2023 IEEE/CVF international conference on computer vision (ICCV)}, 2023, pp. 3992--4003.

\bibitem{pr++}
X.~Yu, Y.~Xie, Y.~Liu, H.~Lu, R.~Xiong, Y.~Liao, and Y.~Wang, ``Leverage cross-attention for end-to-end open-vocabulary panoptic reconstruction,'' \emph{arXiv preprint arXiv:2501.01119}, 2025.

\end{thebibliography}


\end{document}